\documentclass[runningheads]{ASMDA}

\usepackage{graphicx} 

\usepackage{amsmath,amssymb,amsopn}
\usepackage[mathscr]{euscript}
\usepackage{rotating}
\begin{document}
%
\title*{A comparison of bandwidth selectors for mean shift clustering}
%
\toctitle{A comparison of bandwidth selectors for mean shift clustering}
%
\titlerunning{Bandwidth selectors for mean shift clustering}
%
\author{
Jos\'e E. Chac\'on\inst{1},
\and
Pablo Monfort\inst{1}
}
%
\index{Chac\'on, J. E.}
\index{Monfort, P.}

%
\authorrunning{Chac\'on and Monfort}
%
\institute{
Departamento de Matem\'aticas, Universidad de Extremadura, E-06006 Badajoz, Spain\\
(E-mail: {\tt jechacon@unex.es}, {\tt pabmonf@unex.es})
}

\maketitle

\begin{abstract}
We explore the performance of several automatic bandwidth selectors, originally designed for
density gradient estimation, as data-based procedures for nonparametric, modal clustering. The key
tool to obtain a clustering from density gradient estimators is the mean shift algorithm, which
allows to obtain a partition not only of the data sample, but also of the whole space. The results
of our simulation study suggest that most of the methods considered here, like cross validation and
plug in bandwidth selectors, are useful for cluster analysis via the mean shift algorithm.
\end{abstract}

\section{Introduction}

The mean shift algorithm was introduced by Fukunaga and Hostetler in a seminal paper in 1975
\cite{FH75}, with the goal of estimating the gradient of a multivariate density. They also showed
that their algorithm can be helpful for many applications in several pattern recognition problems,
and particularly pointed out its usefulness for clustering and data filtering.

Even if this algorithm was highlighted in the popular book by Silverman \cite{Sil86} as an
important application of kernel smoothing, it remained relatively neglected in the Statistics
literature, until it was ``re-discovered" by Cheng \cite{Ch95}, Carreira-Perpi\~n\'an \cite{CP00}
and Comaniciu and Meer \cite{CM02} for its applications in Engineering. Some recent contributions
that make use of the mean shift algorithm, either explicitly or implicitly, are \cite{LRL07},
 \cite{OE11}, \cite{GPVW12} or \cite{CD13}.

Being closely related to the problem of density gradient estimation, the mean shift algorithm
inherits its dependence on the choice of a suitable bandwidth matrix. It was only recently (see
\cite{CD13} and \cite{HKV13}) that automatic methods for bandwidth selection for density gradient
estimation were proposed. The goal of this paper is to provide a comparative study of the
performance of these automatic bandwidth selectors, not with respect to the problem of density
gradient estimation, but regarding the clustering of the space that they induce via the mean shift
algorithm.

The rest of the paper is organized as follows. In Section 2 below the clustering procedure derived
from the mean shift algorithm is introduced. A brief review of the existing bandwidth matrix
selectors for density gradient estimation is contained in Section 3. The details of the simulation
study comparing these methodologies are given in Section 4 and some conclusions are discussed in
Section 5. Finally, we show in an Appendix the ascending property of the mean shift algorithm with
an unconstrained bandwidth matrix.

\section{Mean shift clustering}

Let us consider a probability density $f\colon\mathbb R^d\to\mathbb R$, and denote by $\mathsf D f$
its gradient vector, so that with the usual column notation for vectors $\boldsymbol
x=(x_1,\dots,x_d)^\top$ we have $$\mathsf D f=\frac{\partial f}{\partial \boldsymbol
x}=\Big(\frac{\partial f}{\partial x_1},\ldots,\frac{\partial f}{\partial x_d}\Big)^\top,$$ with
$^\top$ standing for the transpose operator.

The mean shift algorithm is a variant of the well-known gradient ascent algorithm which is usually
employed to find the local maxima of a given function. Explicitly, given any starting point
$\boldsymbol y_0\in\mathbb R^d$, the mean shift algorithm iteratively constructs a sequence
$(\boldsymbol y_0,\boldsymbol y_1,\boldsymbol y_2,\ldots)$ according to the following updating
mechanism
\begin{equation}\label{eq:meanshift}
\boldsymbol y_{j+1}=\boldsymbol y_j+\mathbf A {\mathsf D f}(\boldsymbol y_j)\big/f(\boldsymbol y_j),
\end{equation}
where $\mathbf A$ is a $d\times d$ positive definite matrix conveniently chosen to guarantee the
convergence of the sequence. The only difference with the usual gradient ascent algorithm is that
(\ref{eq:meanshift}) uses the normalized gradient $\mathsf D f/f$ to accelerate the convergence
when the starting point belongs to a low-density zone.

Since the shift at every step is done approximately along the gradient direction it follows that
the limit point of the mean shift sequence should be a local maximum of $f$ (i.e., a mode of the
density). This induces a clustering scheme in which any two points are said to belong to the same
cluster whenever the sequences constructed from them as starting points converge to the same mode
of $f$. In that case, it is also common to say that the two points belong to the same domain of
attraction of such local maximum, and this type of clustering is called {\it modal clustering}.

Moreover, since the mean shift algorithm is applicable with any point in $\mathbb R^d$ as starting
point, eventually this clustering scheme induces a partition of the whole space $\mathbb R^d$ into
disjoint clusters. This partition, built up from the knowledge of the density $f$, will be referred
to as the ideal population clustering. A precise definition of this ideal population clustering can
be found in \cite{Ch12}.

When the density $f$ is unknown, but a sample $\mathbf X_1,\dots,\mathbf X_n$ from $f$ is observed
instead, the mean shift algorithm (\ref{eq:meanshift}), with the density and the density gradient
estimated from the sample, yields a data-based clustering of the whole space $\mathbb R^d$.

The goal of most clustering methodologies is not to partition $\mathbb R^d$, but only the data
sample. Nevertheless, it is clear that by partitioning the whole space the mean shift algorithm
induces, in particular, a clustering of the data by assigning two data points to the same cluster
if they belong to the same component of the aforementioned partition of $\mathbb R^d$.

The density and density gradient estimators considered here are of kernel type. The kernel density
estimator has the form
$$\hat f_{\mathbf H}(\boldsymbol x)=n^{-1}\sum_{i=1}^nK_{\mathbf H}(\boldsymbol x-\mathbf X_i),$$
where the kernel $K$ is a spherically symmetric $d$-variate density function, the bandwidth matrix
$\mathbf H$ is symmetric and positive definite, and we have used the re-scaling notation
$K_{\mathbf H}(\boldsymbol x)=|\mathbf H|^{-1/2}K(\mathbf H^{-1/2}\boldsymbol x)$ (see \cite{WJ95},
Chapter 4). Then, following \cite{CDW11}, the density gradient estimator is just the gradient of
the kernel density estimator, given by
$$\mathsf D \hat f_{\mathbf H}(\boldsymbol x)=n^{-1}\sum_{i=1}^n\mathsf D K_{\mathbf H}(\boldsymbol x-\mathbf X)=n^{-1}|\mathbf H|^{-1/2}\mathbf H^{-1/2}\sum_{i=1}^n\mathsf D K\big(\mathbf H^{-1/2}(\boldsymbol x-\mathbf X)\big)$$

Being symmetric, the kernel $K$ can be expressed as $K(\boldsymbol x)=\tfrac12k(\boldsymbol
x^\top\boldsymbol x)$, where the function $k\colon\mathbb R_+\to\mathbb R$ is known as the profile
of $K$ (see \cite{CM02}). Furthermore, the kernel $K$ is usually assumed to be smooth and unimodal,
so that $g(x)=-k'(x)\geq0$. Thus, following the ideas of \cite{FH75}, \cite{CD13} showed that a
sensible estimator of the normalized gradient $\mathsf D f(\boldsymbol x)/f(\boldsymbol x)$ is
$\mathbf H^{-1}\mathbf m_{\mathbf H}(\boldsymbol x)$, where the term $\mathbf m_{\mathbf
H}(\boldsymbol x)=\sum_{i=1}^n\omega_{i,\mathbf H}(\boldsymbol x)\mathbf X_i-\boldsymbol x$
is known as the {\em mean shift}. It is the difference between a weighted mean of the data and
$\boldsymbol x$, with the weights $\omega_{i,\mathbf H}(\boldsymbol x)$ defined as
$$\omega_{i,\mathbf H}(\boldsymbol x)=\frac{g\big(M_{\mathbf H}(\boldsymbol x,\mathbf X_i)\big)}{\sum_{\ell=1}^ng\big(M_{\mathbf H}(\boldsymbol x,\mathbf X_\ell)\big)},$$
where $M_{\mathbf H}$ denotes the Mahalanobis distance $M_{\mathbf H}(\boldsymbol x,\boldsymbol
y)=(\boldsymbol x-\boldsymbol y)^\top\mathbf H^{-1}(\boldsymbol x-\boldsymbol y)$. Hence, by
plugging this estimate in (\ref{eq:meanshift}) and taking $\mathbf A=\mathbf H$, the updating
mechanism of the data-based mean shift algorithm simply reads
\begin{equation}\label{eq:ms2}
\boldsymbol y_{j+1}=\sum_{i=1}^n\omega_{i,\mathbf H}(\boldsymbol y_j)\mathbf X_i.
\end{equation}

Originally, \cite{FH75} developed the mean shift algorithm using a constrained bandwidth matrix
consisting of a scalar $h^2$ times the identity matrix, $h>0$, and \cite{CM02} showed that for this
constrained form the choice of $\mathbf A=\mathbf H$ guarantees that the mean shift sequence is
convergent, as long as the kernel $K$ has a convex and monotonically decreasing profile $k$. In the
Appendix below we show that (\ref{eq:ms2}), the unconstrained version of the mean shift algorithm,
is also convergent.

\section{Bandwidth matrix selectors}

As it is common for all kernel smoothing methods, the performance of mean shift clustering is
highly influenced by the choice of the bandwidth matrix. Since the element having the biggest
impact on the performance of the mean shift algorithm appears to be the density gradient, it seems
reasonable that a bandwidth matrix chosen to obtain a good kernel density gradient estimate could
lead to an appealing clustering via the mean shift algorithm.

Surprisingly, the literature tackling the problem of automatic, data-based bandwidth matrix
selection for kernel density gradient estimation is quite scant and recent. We are aware only of
two contributions dealing with this problem: \cite{CD13} and \cite{HKV13}. In both papers the
measure to evaluate the performance of the kernel density gradient estimator $\mathsf D\hat
f_\mathbf H$ is the mean integrated squared error, defined as
$${\rm MISE}(\mathbf H)=\int_{\mathbb R^d}\|\mathsf D\hat
f_\mathbf H(\boldsymbol x)-\mathsf D f(\boldsymbol x)\|^2d\boldsymbol x,$$ where $\|\cdot\|$
denotes the usual Euclidean norm in $\mathbb R^d$. With this goal in mind, the optimal bandwidth
for kernel density gradient estimation is taken to be $\mathbf H_{\rm MISE}$, the minimizer of the
MISE function over the class of all symmetric positive definite matrices.

In \cite{CD13}, three bandwidth matrix selectors were proposed for kernel estimation of the $r$-th
derivative of a multivariate density $f$, for arbitrary $r$. They are defined as the minimizers of
certain criteria which aim to estimate the MISE. These criteria can be shown to generalize the
well-known cross validation (CV), plug-in (PI) and smooth cross validation (SCV) methodologies
proposed earlier for the base case of univariate density estimation (i.e., $d=1$ and $r=0$). In the
case of the density gradient (arbitrary $d$ and $r=1$), these criteria can be written as
\begin{align*}
{\rm CV}(\mathbf H)&=-n^{-2}
\sum_{i,j=1}^n\nabla^2K_{2\mathbf H}(\mathbf X_i-\mathbf X_j)+2[n(n-1)]^{-1}\sum_{i\neq j}\nabla^2 K_\mathbf H(\mathbf X_i-\mathbf X_j)\\
{\rm PI}(\mathbf H)&= n^{-1}|\mathbf H|^{-1/2}{\rm tr}\big\{\mathbf H^{-1}\mathbf R(\mathsf DK)\big\}\\&\quad-
\tfrac{1}{4} \{({\rm vec}^\top\mathbf I_{d}) \otimes ({\rm vec}^\top \mathbf H) \otimes ({\rm vec}^\top \mathbf H)\} n^{-2} \sum_{i,j=1}^n \mathsf D^{\otimes 6} K_\mathbf G (\mathbf X_i - \mathbf X_j)\\
{\rm SCV}(\mathbf H)&=n^{-1}|\mathbf H|^{-1/2}{\rm tr}\big\{\mathbf H^{-1}\mathbf R(\mathsf DK)\big\} \\
&\quad-n^{-2}\sum_{i,j=1}^n\nabla^2\big\{K_{2\mathbf H+2\mathbf G}-2K_{\mathbf H+2\mathbf G}+K_{2\mathbf G}\big\}(\mathbf X_i-\mathbf X_j),
\end{align*}
respectively. Here, $\nabla^2=\sum_{i=1}^d(\partial^2/\partial x_i^2)$ is the Laplace operator,
${\rm tr}$ denotes the trace operator, ${\rm vec}$ is the vectorization operator that transforms a
matrix into a vector by stacking the columns of the matrix one underneath the other, $\otimes$
denotes the Kronecker product, $\mathbf R(\mathsf DK) = \int_{\mathbb R^d} \mathsf DK(\boldsymbol
x) \mathsf DK(\boldsymbol x)^\top d\boldsymbol x$ is a $d\times d$ matrix, the vector $\mathsf
D^{\otimes 6} K_\mathbf G\in\mathbb R^{d^6}$ includes all 6-th order partial derivatives of
$K_\mathbf G$, arranged in a particular order (see \cite{CD10}), and $\mathbf G$ is a pilot
bandwidth matrix. Computation of these criteria is not simple, but efficient implementations were
proposed in \cite{CD13b}.

In \cite{HKV13} an iterative method (IT) was proposed to treat the cases $r=0$ and $r=1$ for
arbitrary $d$. These authors noted that the asymptotic approximation of the optimal bandwidth
$\mathbf H_{\rm MISE}$, so-called $\mathbf H_{\rm AMISE}$, can be characterized as the solution of
a particular equation involving the unknown density $f$. So a sensible choice for the bandwidth is
introduced as the solution of a data-based estimate of this equation, which for $r=1$ can be
written as
\begin{multline*}
(d+2)n^{-1}|\mathbf H|^{-1/2}{\rm tr}\big\{\mathbf H^{-1}\mathbf R(\mathsf DK)\big\}\\
+4n^{-2}\sum_{i,j=1}^n\nabla^2\big\{K_{2\mathbf H+2\mathbf G}-2K_{\mathbf H+2\mathbf G}+K_{2\mathbf
G}\big\}(\mathbf X_i-\mathbf X_j)=0.
\end{multline*}
Again, the computational details to obtain the solution of this equation are not simple, and an
iterative method to solve it (hence the name of this bandwidth selector) is proposed in
\cite{HKV13}.

All these methodologies focus on the most general form for the bandwidth matrix $\mathbf H$, which
is only required to be symmetric and positive definite. Other popular choices for the bandwidth
matrix include constrained forms such as $\mathbf H$ being diagonal, $\mathbf H={\rm
diag}(h_1^2,\dots,h_d^2)$, or the parametrization using a single bandwidth $h>0$ so that $\mathbf
H=h^2\mathbf I_d$, with $\mathbf I_d$ denoting the $d\times d$ identity matrix.

The thorough study of \cite{WJ93} reported that for density estimation, in general, the diagonal
parametrization results in a small loss of efficiency, but the single-bandwidth estimator should
not be blindly used for unscaled multivariate data (see also \cite{Ch09}). For density derivative
estimation, \cite{CDW11} showed that the loss of efficiency due to the use of simpler bandwidth
matrix parametrizations can be even more severe. However, the goal of cluster analysis is quite
different from that of density estimation, so that not very precise density estimates may equally
lead to nearly optimal clusterings (see \cite{Ch12}, Figure 6, for an illustration of this
phenomenon), so in principle the simpler parametrizations should not be completely discarded. In
fact, the very simple diagonal bandwidth proposal of \cite{AT07} was shown to produce good results
in \cite{CD13}. Therefore, unconstrained but also diagonal bandwidth matrices will be considered in the simulation study below.

\section{Simulation study}

The main goal of this paper is to provide an empirical comparison of the performance of several
bandwidth selection methods for mean shift clustering.

\begin{figure}
\centering
\begin{tabular}{@{}cc@{}}
Trimodal III& Quadrimodal\\
    \includegraphics[width=0.5\textwidth]{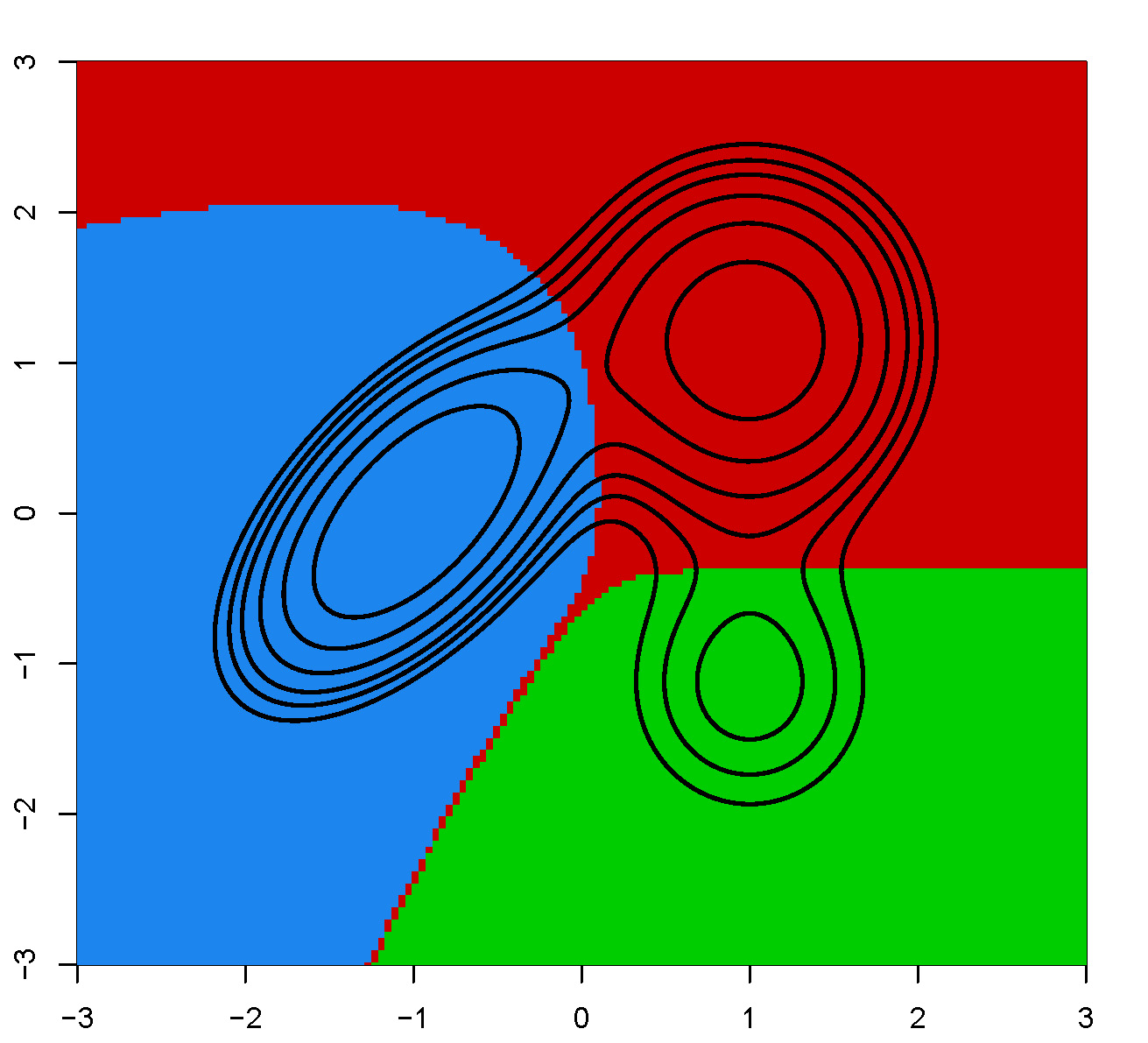}&\includegraphics[width=0.5\textwidth]{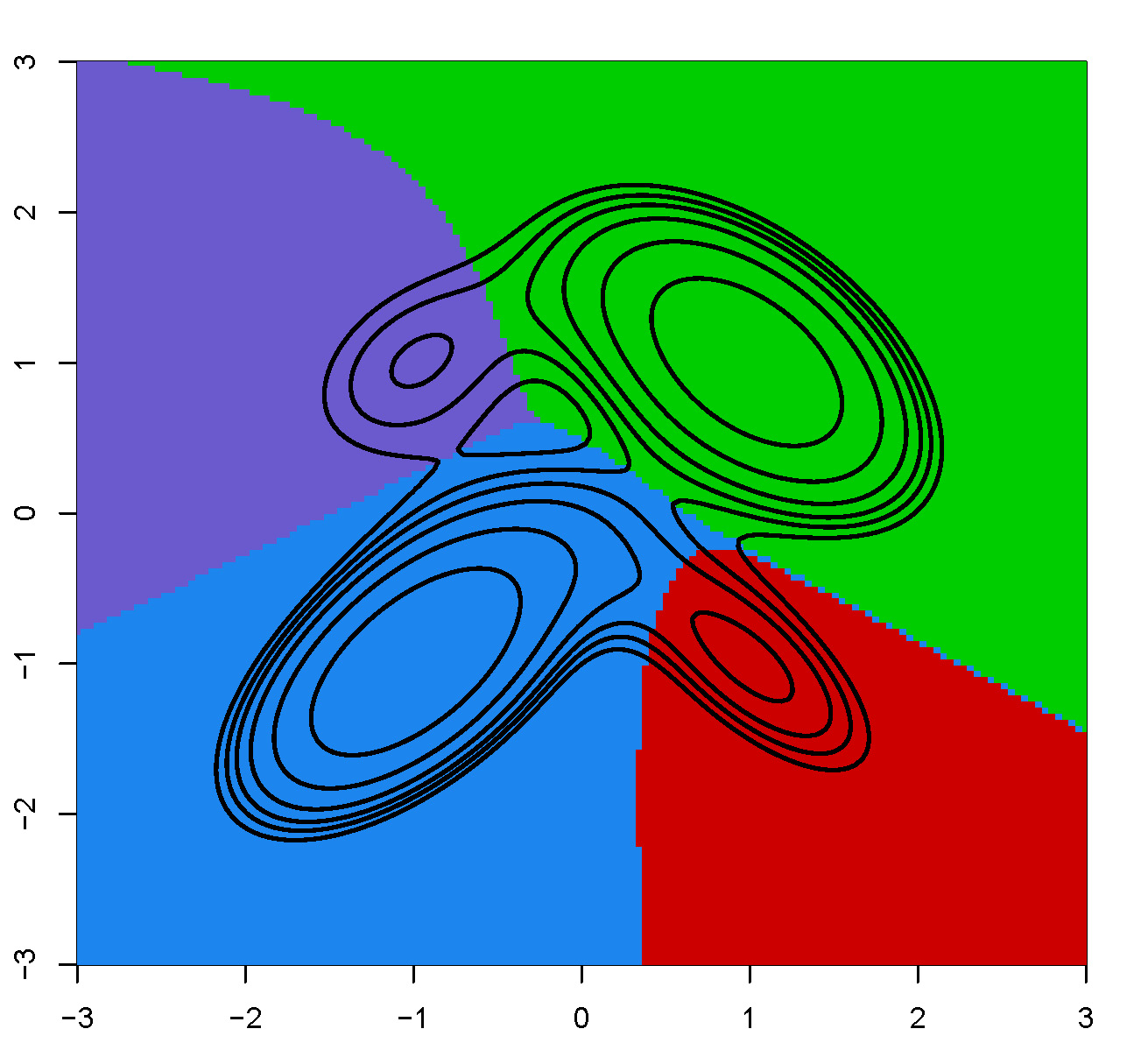}\\[10pt]

4-crescent& Broken ring\\

  \includegraphics[width=0.5\textwidth]{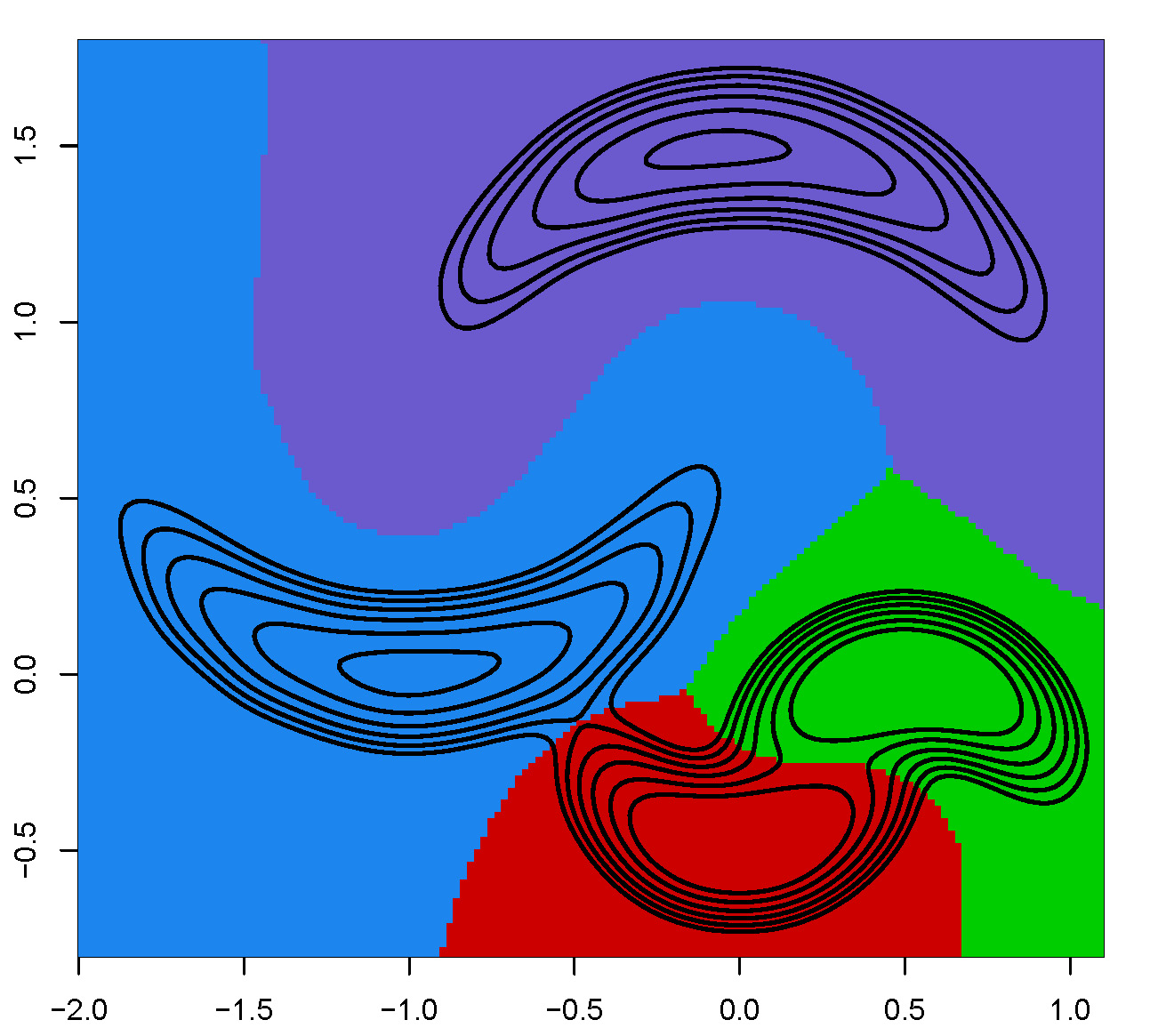}&\includegraphics[width=0.5\textwidth]{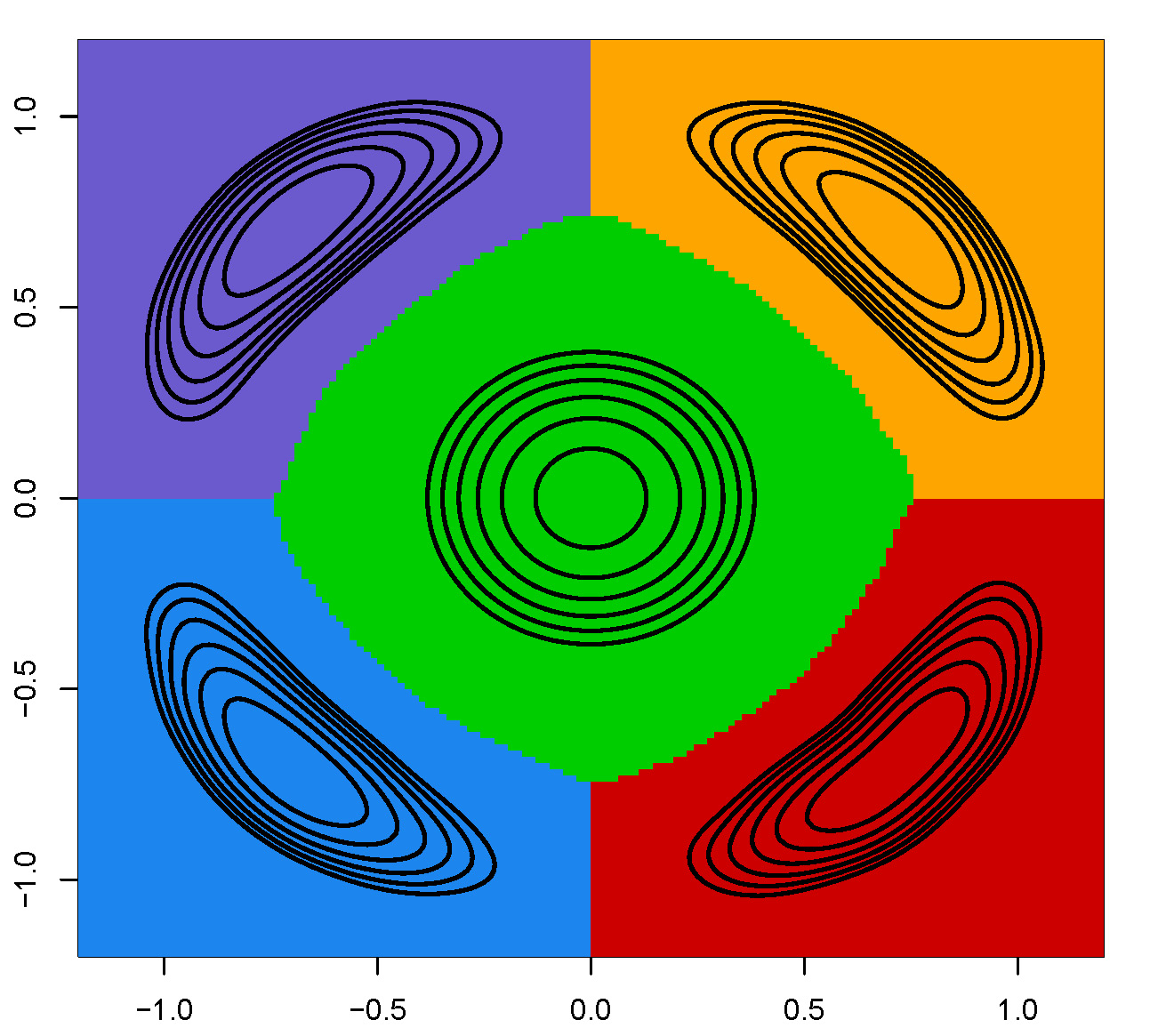}\\[10pt]
\end{tabular}
\begin{tabular}{@{}c@{}}
Eye\\
  \includegraphics[width=0.5\textwidth]{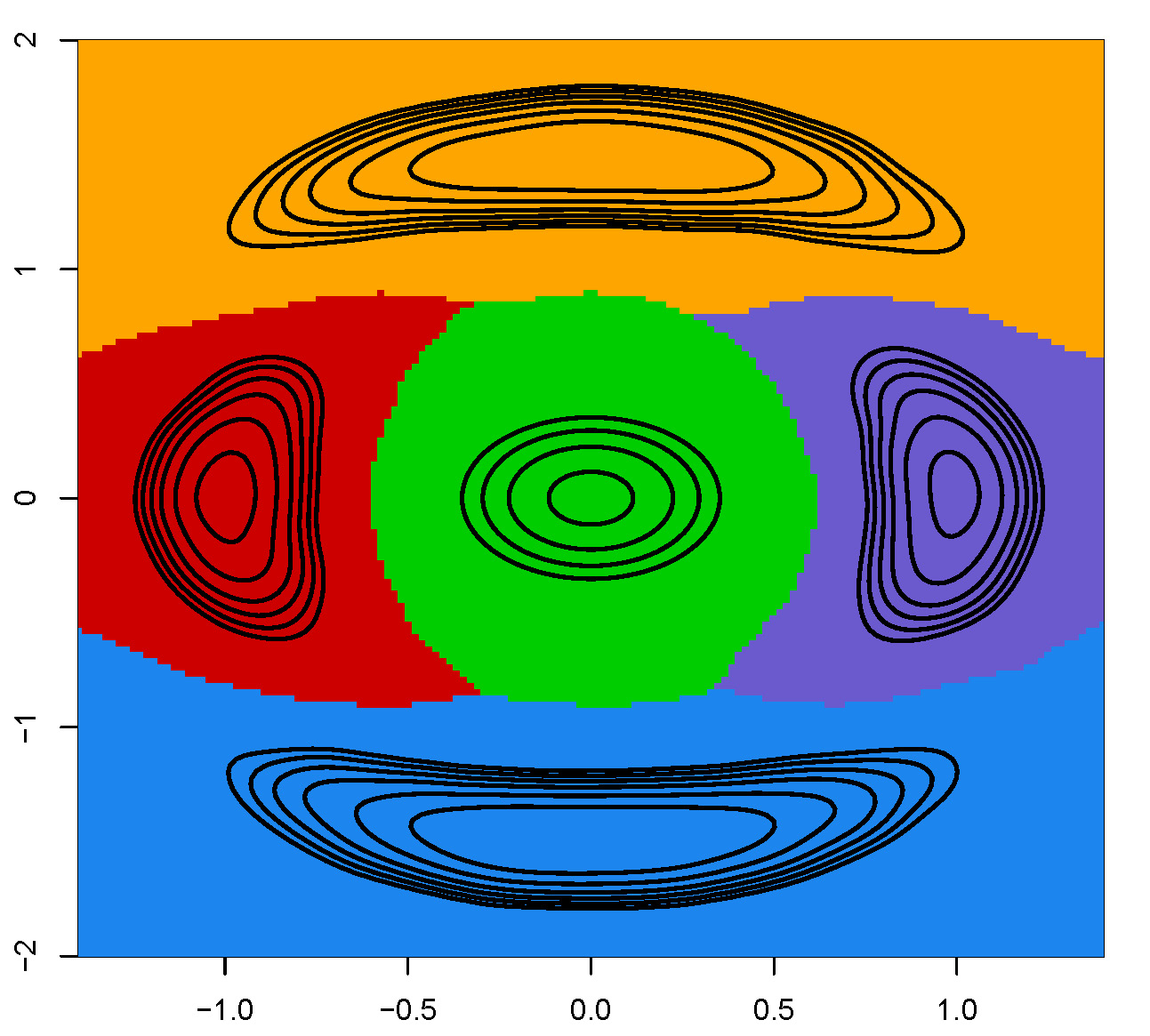}
\end{tabular}
  \caption{The five true density models included in the simulation study, with the ideal population clustering shown in different colors.}
  \label{fig:1}
\end{figure}

Five true models are analyzed in the study, which cover a wide variety of cluster shapes. Two of these
densities are normal mixture densities; hence a parametric cluster analysis of these two models, by fitting an estimated density through maximum likelihood, would probably yield quite good results (see \cite{CD13}, and references therein). But to exploit the nonparametric nature of the mean shift approach we also include
three densities with more intricate features which are not likely to be accurately recovered in a parametric
setup. Figure \ref{fig:1} shows the true densities and the ideal population clusterings associated
to each of these models, along with the names which we will use to refer to them. A precise definition of these models can be found in \cite{CD13}.

The automatic bandwidth selectors compared in this study are the CV, PI and SCV bandwidths proposed
in \cite{CD13} for density gradient estimation, the IT method introduced in \cite{HKV13}, the normal-scale bandwidth (NS) for density gradient estimation introduced in \cite{CDW11}, and the simple proposal AT of \cite{AT07}, which shrinks the
 diagonal normal-scale bandwidth for density estimation by a factor $3/4$ (hence, it could be considered as a diagonal variant of the previous one). For the CV, PI, SCV and IT methods we also considered their respective diagonal versions, which are obtained by minimizing (or solving, in the case of IT) the objective criteria over the class of all positive definite diagonal matrices. These are denoted by adding `D' to
their initials (i.e., CVD, PID, SCVD and ITD), while their unconstrained counterparts are
represented by CVU, PIU, SCVU and ITU, respectively.

The measure of the performance of each of these methods is completely different than that employed
in \cite{CD13}. There, different clusterings of the data were compared by means of the adjusted
Rand index criterion, introduced in \cite{HA85}. Here, the interest is not to compare different
clusterings of the data, but clusterings of the whole space $\mathbb R^d$. Therefore, it is
necessary to use a distance between clusterings of $\mathbb R^d$, and we will use the {\it distance in
measure} proposed in \cite{Ch12}.

This distance is defined as follows: given two clusterings $\mathscr C=\{C_1,\dots,C_r\}$ and $\mathscr D=\{D_1,\dots,D_s\}$ of a probability measure $P$, with $r\leq s$, the distance in measure between them is defined as
$$d_P(\mathscr C,\mathscr D)=\frac12\min_{\sigma\in \mathcal P_s}\sum_{i=1}^sP(C_i\triangle D_{\sigma(i)}),
$$
where $\mathcal P_s$ denotes the set of permutations of $\{1,2,\ldots,s\}$, the partition $\mathscr
C$ has been enlarged by adding $s-r$ empty sets $C_{r+1}=\cdots=C_s=\varnothing$ if necessary, and
$\triangle$ denotes the symmetric difference between two sets, namely $C\triangle D=(C\cap
D^c)\cup(C^c\cap D)$.

Even if the true densities are known in a simulation study, the exact value of $d_P(\mathscr
C,\mathscr D)$ is difficult to compute in practice. We used a fine enough grid defined over a large
rectangle chosen to contain at least 0.999 probability mass. This regular grid is ruled in
rectangles by considering a tiny rectangle centered at each grid point with its sides of length
half the distance to the next grid point in each coordinate direction. Each grid point is assigned
to one cluster via the mean shift algorithm, and hence every cluster can be approximated by the
union of tiny rectangles surrounding the grid points that are labeled to belong to it. By computing
the probability mass of each tiny rectangle and adding up the contributions corresponding to the
rectangles that approximate each symmetric difference we obtain an approximation of $P(C_i\triangle
D_j)$ for each $i,j=1,\dots,s$. Finally, finding the minimum over all the permutations $\mathcal
P_s$ is known as a linear sum assignment problem, and efficient algorithms to solve it are shown,
e.g., in \cite{PS82}.

One hundred samples of size $n=500$ from each density in the study were drawn. For each of these
samples all the ten bandwidth selectors NS, AT, CVU, CVD, PIU, PID, SCVU, SCVD, ITU and ITD were
computed and a partition in clusters of the whole space was obtained through the mean shift
algorithm. Finally, we recorded the distance in measure between such data-based partitions and the
ideal population clustering. The sample medians and interquartile ranges (IQR) of these distances
over the 100 samples are summarized in Table \ref{tab:1}. The median and the IQR were preferable
over the more usual mean and standard deviation because the distribution of these random distances
in measure was generally skewed and contained some outliers.

\begin{sidewaystable}[p!t]
\vspace{11cm}\centering
\begin{tabular}{l||cccccccccc}
\multicolumn{1}{c||}{\quad}  &\multicolumn{10}{c}{\sc Bandwidth selector}\\
\multicolumn{1}{c||}{\sc Model}        & NR & AT & CVU & CVD & PIU & PID & SCVU & SCVD & ITU & ITD\\
\hline\hline

Trimodal III
&    6.37e-02 & 6.37e-02 & 1.04e-03 & 2.05e-04 & 9.96e-05 & 9.26e-05 & 3.19e-02 & 6.37e-02 & {\bf 7.02e-05} & 1.04e-04\\
& (6.36e-02) & (5.15e-06) & (5.29e-02) & (3.42e-02) & (4.60e-03) & (1.27e-03) & (6.36e-02) & (6.36e-02) & (1.98e-02) & (5.19e-02) \\
       \hline
Quadrimodal
&    9.70e-02 & 9.68e-02 & {\bf 1.63e-02} & 4.75e-02 & 3.16e-02 & 4.74e-02 & 9.70e-02 & 9.68e-02 & 5.24e-02 & 9.79e-02 \\
&  (2.34e-04) & (1.34e-04) & (4.35e-02) & (4.59e-02) & (4.65e-02) & (4.77e-02) & (3.09e-04) & (2.13e-04) & (4.82e-02) & (2.05e-01)\\

       \hline
4-crescent
&  1.21e-01 & 2.44e-01 & 2.03e-21 & 2.24e-21 & 2.44e-03 & 3.44e-03 & {\bf 1.90e-21} & {\bf 1.84e-21} & 1.21e-01 & 3.70e-01  \\
&  (0.00e+00) & (0.00e+00) & (5.16e-22) & (5.34e-04) & (1.49e-02) & (3.91e-02) & (2.42e-05) & (5.51e-22) & (1.23e-01) & (0.00e+00) \\
       \hline
Broken ring
&    3.62e-14 & 2.81e-01 & 3.72e-14 & 3.36e-14 & {\bf 2.58e-14} & {\bf 2.53e-14} & 3.20e-14 & 3.12e-14 & 3.77e-01 & 3.77e-01 \\
& (6.12e-15) & (9.57e-02) & (9.90e-15) & (4.81e-15) & (5.33e-15) & (6.12e-15) & (4.59e-15) & (4.19e-15) & (0.00e+00) & (0.00e+00) \\
       \hline
Eye
&    2.67e-02 & 2.67e-02 & 1.61e-15 & 2.44e-02 & {\bf 3.24e-16} & {\bf 3.37e-16} & 4.40e-16 & 4.50e-16 & 3.24e-01 & 3.24e-01 \\
&  (4.51e-17) & (1.04e-17) & (2.67e-02) & (2.67e-02) & (7.03e-17) & (3.20e-04) & (2.54e-17) & (3.34e-17) & (0.00e+00) & (0.00e+00) \\
       \hline
\end{tabular}
\caption{Sample median and (interquartile range) for the distance in measure between the data-based clusterings induces by each bandwidth selection method and the ideal population clustering along 100 simulation runs of sample size $n=500$ of each distribution. The significantly best methods for each model are marked in bold font.} 
\label{tab:1}
\end{sidewaystable}

\begin{table}[p]
\centering
\begin{tabular}{@{}c @{\hspace{0.2\textwidth}}c@{}}
Trimodal & Quadrimodal\\[4pt]
\begin{tabular}{l|rrrrrrr}
& \multicolumn{7}{c}{No. of clusters}\\
  \cline{2-8}
${\mathbf H}$ & 1 & 2 & {\bf 3} & 4 & 5 & 6 & 7 \\
  \hline
NR &   1 &  54 &  43 &   2 &   0 &   0 &   0 \\
  AT &   1 &  90 &   8 &   0 &   0 &   0 &   0 \\
  CVU &   1 &  11 &  34 &  28 &  12 &   9 &   1 \\
  CVD &   0 &  13 &  41 &  24 &  12 &   2 &   3 \\
  PIU &   1 &  14 &  51 &  23 &  10 &   1 &   0 \\
  PID &   0 &  10 &  48 &  31 &   9 &   0 &   0 \\
  SCVU &   4 &  46 &  45 &   5 &   0 &   0 &   0 \\
  SCVD &   1 &  50 &  45 &   3 &   0 &   0 &   0 \\
  ITU &   2 &  20 &  57 &  18 &   3 &   0 &   0 \\
  ITD &   1 &  19 &  62 &  14 &   2 &   0 &   0 \\
   \hline
\end{tabular}&

\begin{tabular}{l|rrrrrrr}
& \multicolumn{7}{c}{No. of clusters}\\
  \cline{2-8}
$\mathbf H$ & 1 & 2 & 3 & {\bf 4} & 5 & 6 & 7 \\
  \hline
NR &   0 &  87 &  13 &   0 &   0 &   0 &   0 \\
  AT &   0 &  97 &   3 &   0 &   0 &   0 &   0 \\
  CVU &   0 &  14 &  28 &  31 &  16 &   7 &   2 \\
  CVD &   0 &  19 &  44 &  27 &   8 &   1 &   0 \\
  PIU &   0 &  20 &  30 &  30 &  13 &   5 &   0 \\
  PID &   0 &  18 &  36 &  31 &  12 &   1 &   0 \\
  SCVU &   0 &  83 &  16 &   1 &   0 &   0 &   0 \\
  SCVD &   0 &  86 &  14 &   0 &   0 &   0 &   0 \\
  ITU &   0 &  43 &  39 &  13 &   3 &   1 &   0 \\
  ITD &  38 &  50 &  10 &   2 &   0 &   0 &   0 \\
   \hline
\end{tabular}

\\[80pt]
4-crescent & Broken ring\\[4pt]

\begin{tabular}{l|rrrrrrrrr}
& \multicolumn{9}{c}{No. of clusters}\\
  \cline{2-10}
$\mathbf H$ & 1 & 2 & 3 & {\bf 4} & 5 & 6 & 7 & 8 & 9 \\
  \hline
NR &   0 &   8 &  82 &   7 &   2 &   0 &   1 &   0 &   0 \\
  AT &   0 &  99 &   1 &   0 &   0 &   0 &   0 &   0 &   0 \\
  CVU &   0 &   0 &   1 &  81 &  18 &   0 &   0 &   0 &   0 \\
  CVD &  15 &   1 &   0 &  63 &  17 &   0 &   3 &   0 &   0 \\
  PIU &   0 &   0 &   0 &  13 &  28 &  37 &  18 &   2 &   2 \\
  PID &   0 &   0 &   0 &  12 &  23 &  43 &  14 &   5 &   3 \\
  SCVU &   0 &   0 &   0 &  74 &  24 &   2 &   0 &   0 &   0 \\
  SCVD &   0 &   0 &   0 &  76 &  21 &   2 &   1 &   0 &   0 \\
  ITU &  22 &  16 &  45 &   9 &   4 &   2 &   0 &   0 &   0 \\
  ITD & 100 &   0 &   0 &   0 &   0 &   0 &   0 &   0 &   0 \\
   \hline
\end{tabular}

&\begin{tabular}{l|rrrrrrr}
& \multicolumn{7}{c}{No. of clusters}\\
  \cline{2-8}
$\mathbf H$ & 1 & 2 & 3 & 4 & {\bf 5} & 6 & 7 \\
  \hline
NR &   0 &   0 &   0 &   0 & 100 &   0 &   0 \\
  AT &  48 &  33 &  12 &   6 &   1 &   0 &   0  \\
  CVU &   0 &   0 &   9 &   2 &  86 &   2 &   1  \\
  CVD &   0 &   0 &   0 &   0 &  98 &   1 &   0  \\
  PIU &   0 &   0 &   0 &   0 &  95 &   5 &   0  \\
  PID &   0 &   0 &   0 &   0 &  92 &   7 &   1  \\
  SCVU &   0 &   0 &   0 &   0 &  98 &   2 &   0 \\
  SCVD &   0 &   0 &   0 &   0 & 100 &   0 &   0 \\
  ITU & 100 &   0 &   0 &   0 &   0 &   0 &   0  \\
  ITD & 100 &   0 &   0 &   0 &   0 &   0 &   0  \\
   \hline
\end{tabular}\\[80pt]
\end{tabular}

\begin{tabular}{@{}c@{}}
Eye\\[4pt]
\begin{tabular}{l|rrrrrrr}
& \multicolumn{7}{c}{No. of clusters}\\
  \cline{2-8}
$\mathbf H$ & 1 & 2 & 3 & 4 & {\bf 5} & 6 & 7 \\
  \hline
NR &   0 &   0 &   0 & 100 &   0 &   0 &   0 \\
  AT &   0 &   0 &   1 &  98 &   1 &   0 &   0 \\
  CVU &   0 &   0 &   0 &  44 &  52 &   3 &   1 \\
  CVD &   0 &   0 &   0 &  30 &  42 &  12 &   3 \\
  PIU &   0 &   0 &   0 &   0 &  78 &  18 &   3 \\
  PID &   0 &   0 &   0 &   0 &  67 &  28 &   4 \\
  SCVU &   0 &   0 &   0 &   0 &  99 &   1 &   0 \\
  SCVD &   0 &   0 &   0 &   0 &  99 &   1 &   0 \\
  ITU &  81 &   8 &   8 &   3 &   0 &   0 &   0 \\
  ITD & 100 &   0 &   0 &   0 &   0 &   0 &   0 \\
   \hline
\end{tabular}
\end{tabular}

\caption{Distribution of the number of clusters for each clustering method along the five density
models. The true number of clusters is marked in bold font.} \label{tab:2}
\end{table}

In view of Table \ref{tab:1} it is clear that no bandwidth selector is uniformly preferable over
the others. However, it seems clear that NR and AT nearly always induced a poor clustering (an
exception is NR for the broken ring model). The reason for this bad performance could be partially
explained by Table \ref{tab:2}. There it is shown the distribution of the number of clusters
obtained by each method along the 100 simulation runs for each density model (some unusually large
number of clusters have been omitted for clarity). In Table \ref{tab:2} it is possible to
appreciate that both NR and AT normally induce a number of clusters smaller than those appearing in
the true model, which can be interpreted as a well-known oversmoothing effect, due to the fact that
 these two bandwidth selectors are  based on a normal reference rule. As noted before, an
exception is the broken ring model, where NR correctly identifies 5 clusters in all the cases. We
acknowledge, however, that the density clustering approach of Azzalini and Torelli \cite{AT07} is
not based on the mean shift algorithm; the goal here was to test if their appealingly simple
bandwidth proposal were also suitable for mean shift clustering.

The performance of the IT methods is somehow erratic. ITU is the best method for the trimodal
mixture density, and has moderately good results for the quadrimodal mixture density as well, but
both ITU and ITD are unable to deal with more complicated features like those appearing in the last
three density models. Again, a partial explanation for this is provided in Table \ref{tab:2}, where
it is shown that ITU, and especially ITD, tend to partition the space in only one cluster, thus
presenting a highly oversmoothed estimate. The fact that both methods found only one cluster in all
the cases for the broken ring model is the reason why the IQR of the distribution of their
distances in measure is exactly zero (the distance in measure is a constant variable in this case).

The PI bandwidth selectors induce the clusterings with lowest distance in measure for the broken
ring and eye models, and are close to the best performance in the two normal mixture density
models, ranking second to best in terms of median error. They fail, however, to capture the
features of the 4-crescent density, with a tendency to find more clusters than present (as seen in
Table \ref{tab:2}). The CV bandwidths perform disappointingly in the case of the trimodal mixture
density, but CVU ranks first for the quadrimodal density model and both CVU and CVD obtain
moderately good results for the densities with complicated features, frequently finding the right
number of clusters. Both SCV proposals are probably the best ones concerning the right number of
clusters for the densities with complicated features, and indeed they have the best marks for the
4-crescent model, but their behaviour is far from optimal for the normal mixture densities, with
performances close to NR and AT.

With respect to the unconstrained-diagonal bandwidth dilemma, our study seems to suggest that
diagonal bandwidths perform worse than their unconstrained counterparts in most of the cases.
However, perhaps the use of diagonal matrices should not be blindly discarded, since indeed in some
cases their performance is comparable or even slightly better than that of the unconstrained ones, but with a clearly smaller
computational cost.

\section{Conclusion}

We explored here the influence of the bandwidth matrix in the mean shift algorithm from the point
of view of modal clustering. Due to the crucial influence of the density gradient estimate in the
mean shift algorithm we analyzed the practical performance of ten bandwidth selectors originally
designed for density gradient estimation.

None of the ten automatic bandwidth matrix selectors showed a consistent superior performance over
the rest of the methods in our simulation study, but surely neither NR nor AT can be recommended
for general use. All the CV, PI, SCV and IT proposals are best for one of the models, but utterly fail
to identify the cluster structure for one, two or even three of the remaining ones. This
suggests that the problem of bandwidth selection for mean shift clustering, though related, is
different from that of bandwidth selection for density gradient estimation, and presents its own
peculiarities, which undoubtedly deserve to be studied in further detail.

Since CVU and PIU are the only methods that failed solely for one of the density models, any of
these two bandwidth matrix selectors would represent a cautious recommendation in practice, out of the ten
methods studied here.

\bigskip

\noindent{\bf Acknowledgements.} We would like to express our gratitude to Prof. Ivana Horov\'a for
encouraging us to write this contribution, and to Dr. Kamila Vopatov\'a for sharing their code for
the iterative bandwidth computation. The first author wishes to extend thanks to the whole research
group leaded by Prof. Horov\'a in Brno (Czech Republic) for their hospitality during his visit to
the Masaryk University. This work has been supported by grant MTM2010-16660 from the Spanish
Ministerio de Ciencia e Innovaci\'on.

\section*{Appendix}


Here it is shown that when the profile of the kernel is a bounded, convex, non-increasing,
differentiable function, then the mean shift is an ascending algorithm; that is, the points of the
sequence $(\boldsymbol y_0,\boldsymbol y_1,\boldsymbol y_2,\ldots)$ obtained through the mean shift
algorithm attain sequentially increasing values of the estimated density $\hat f_{\mathbf H}$, so
that the sequence $\big(\hat f_\mathbf H(\boldsymbol y_0),\hat f_\mathbf H(\boldsymbol y_1),\hat
f_\mathbf H(\boldsymbol y_2),\ldots\big)$ is convergent.

\begin{proof}
Notice that since $K(\boldsymbol x)=\frac12 k(\boldsymbol x^\top\boldsymbol x)$ it follows that
$2|\mathbf H|^{1/2}K_{\mathbf H}(\boldsymbol x-\mathbf X_i)=k\big(M_{\mathbf H}(\boldsymbol
x,\mathbf X_i)\big)$. Therefore,
$$2n|\mathbf H|^{1/2}\big\{\hat f_\mathbf H(\boldsymbol y_{j+1})-\hat f_\mathbf H(\boldsymbol
y_{j})\big\}=\sum_{i=1}^n\big\{k\big(M_\mathbf H(\boldsymbol y_{j+1},\mathbf
X_i)\big)-k\big(M_\mathbf H(\boldsymbol y_{j},\mathbf X_i)\big)\big\}.$$ Then, following \cite{CM02},
the convexity of the profile $k$ implies that
\begin{multline*}
k\big(M_\mathbf H(\boldsymbol y_{j+1},\mathbf
X_i)\big)-k\big(M_\mathbf H(\boldsymbol y_{j},\mathbf X_i)\big)\\\geq k'\big(M_\mathbf H(\boldsymbol
y_{j},\mathbf X_i)\big)\big\{M_\mathbf H(\boldsymbol y_{j+1},\mathbf X_i)-M_\mathbf H(\boldsymbol
y_{j},\mathbf X_i)\big\}.
\end{multline*}
Hence, expanding the difference between the two Mahalanobis distances we
obtain
\begin{multline*}
2n|\mathbf H|^{1/2}\big\{\hat f_\mathbf H(\boldsymbol y_{j+1})-\hat f_\mathbf H(\boldsymbol
y_{j})\big\}\geq-\sum_{i=1}^ng\big(M_\mathbf H(\boldsymbol y_{j},\mathbf
X_i)\big)\\\times\big\{\boldsymbol y_{j+1}^\top\mathbf H^{-1}\boldsymbol y_{j+1}-\boldsymbol
y_{j}^\top\mathbf H^{-1}\boldsymbol y_{j} -2(\boldsymbol y_{j+1}-\boldsymbol y_j)^\top\mathbf
H^{-1}\mathbf X_i\big\}.
\end{multline*}
But definition (\ref{eq:ms2}) of the updating step entails that $\sum_{i=1}^ng\big(M_\mathbf
H(\boldsymbol y_{j},\mathbf X_i)\big)\mathbf X_i=\sum_{i=1}^ng\big(M_\mathbf H(\boldsymbol
y_{j},\mathbf X_i)\big)\boldsymbol y_{j+1}$ so that it is possible to replace $\mathbf X_i$ for
$\boldsymbol y_{j+1}$ in the last term of the previous display, and simplify to get
$$
2n|\mathbf H|^{1/2}\big\{\hat f_\mathbf H(\boldsymbol y_{j+1})-\hat f_\mathbf H(\boldsymbol
y_{j})\big\}\geq\sum_{i=1}^ng\big(M_\mathbf H(\boldsymbol y_{j},\mathbf X_i)\big)M_\mathbf
H(\boldsymbol y_j,\boldsymbol y_{j+1})\geq0,
$$
so the sequence $\big(\hat f_\mathbf H(\boldsymbol y_0),\hat f_\mathbf H(\boldsymbol y_1),\hat
f_\mathbf H(\boldsymbol y_2),\ldots\big)$ is non-decreasing and bounded, hence convergent.
\end{proof}

\end{document}